\documentstyle[aclap,graphics]{article}
\newcommand{\hidden}[1]{}

\author{Ted Briscoe \\ {\tt ejb@cl.cam.ac.uk}\\
Computer Laboratory \\ University of Cambridge \\ Pembroke Street \\
Cambridge CB2 3QG, UK}
\title{\vspace{-0.5in}Co-evolution of Language and of the Language Acquisition Device}
\begin{document}
\maketitle
\vspace{-0.5in}
\begin{abstract}
  A new account of parameter setting during grammatical acquisition is
  presented in terms of Generalized Categorial Grammar embedded in a
  default inheritance hierarchy, providing a natural partial ordering
  on the setting of parameters. Experiments show that several
  experimentally effective learners can be defined in this framework.
  Evolutionary simulations suggest that a learner with default initial
  settings for parameters will emerge, provided that learning is memory
  limited and the environment of linguistic adaptation contains an
  appropriate language.
\end{abstract}

\section{Theoretical Background}

Grammatical acquisition proceeds on the basis of a partial genotypic
specification of (universal) grammar (UG) complemented with a learning
procedure enabling the child to complete this specification
appropriately. The parameter setting framework of Chomsky (1981)
claims that learning involves fixing the values of a finite set of
finite-valued parameters to select a single fully-specified grammar
from within the space defined by the genotypic specification of UG.
Formal accounts of parameter setting have been developed for small
fragments but even in these search spaces contain local maxima and
subset-superset relations which may cause a learner to converge to an
incorrect grammar (Clark, 1992; Gibson and Wexler, 1994; Niyogi and
Berwick, 1995). The solution to these problems involves defining
default, unmarked initial values for (some) parameters and/or ordering
the setting of parameters during learning.

Bickerton (1984) argues for the Bioprogram Hypothesis as an
explanation for universal similarities between historically unrelated
creoles, and for the rapid increase in grammatical complexity
accompanying the transition from pidgin to creole languages.  From the
perspective of the parameters framework, the Bioprogram Hypothesis
claims that children are endowed genetically with a UG which, by
default, specifies the stereotypical core creole grammar, with
right-branching syntax and subject-verb-object order, as in
Saramaccan. Others working within the parameters framework have
proposed unmarked, default parameters (e.g. Lightfoot, 1991), but the
Bioprogram Hypothesis can be interpreted as towards one end of a
continuum of proposals ranging from all parameters initially unset to
all set to default values.

\section{The Language Acquisition Device}

A model of the Language Acquisition Device (LAD) incorporates a UG
with associated parameters, a parser, and an algorithm for updating
initial parameter settings on parse failure during learning.

\subsection{The Grammar (set)}

Basic categorial grammar (CG) uses one rule of application which
combines a functor category (containing a slash) with an argument
category to form a derived category (with one less slashed argument
category). Grammatical constraints of order and agreement are captured
by only allowing directed application to adjacent matching categories.
Generalized Categorial Grammar (GCG) extends CG with further rule
schemata.\footnote{Wood (1993) is a general introduction to Categorial
  Grammar and extensions to the basic theory. The most closely related
  theories to that presented here are those of Steedman (e.g. 1988)
  and Hoffman (1995).} The rules of FA, BA, generalized weak
permutation (P) and backward and forward composition (FC, BC) are
given in Figure~\ref{rules} (where X, Y and Z are category variables,
$\mid$ is a variable over slash and backslash, and $\ldots$ denotes
zero or more further functor arguments). Once permutation is included,
several semantically equivalent derivations for {\it Kim loves Sandy}
become available, Figure~\ref{p-ba} shows the non-conventional
left-branching one. Composition also allows alternative
non-conventional semantically equivalent (left-branching) derivations.
\begin{figure*}
\begin{center}
\begin{tabular}{|ll|}
\hline
\hline
\multicolumn{2}{|c|}{Forward Application:} \\[0.2cm]
X/Y Y $\Rightarrow$\ X & $\lambda$ y [X(y)] (y) $\Rightarrow$\ X(y)\\[0.3cm]
\multicolumn{2}{|c|}{Backward Application:} \\[0.2cm]
Y X$\backslash$Y $\Rightarrow$\ X & $\lambda$ y [X(y)] (y) $\Rightarrow$\ X(y)\\[0.2cm]
\hline
\hline 
\multicolumn{2}{|c|}{Forward Composition:} \\[0.2cm]
X/Y Y/Z $\Rightarrow$\ X/Z & $\lambda$ y [X(y)] $\lambda$ z [Y(z)] $\Rightarrow$\ $\lambda$ z [X(Y(z))]\\[0.2cm]
\multicolumn{2}{|c|}{Backward Composition:} \\[0.2cm]
Y$\backslash$Z X$\backslash$Y $\Rightarrow$\ X$\backslash$Z & $\lambda$ z [Y(z)] $\lambda$ y [X(y)] $\Rightarrow$\ $\lambda$ z [X(Y(z))]\\[0.2cm] 
\hline
\hline
\multicolumn{2}{|c|}{(Generalized Weak) Permutation:} \\[0.2cm]
(X$\mid$Y$_{1}$)$\ldots\mid$Y$_{n}$ $\Rightarrow$\
(X$\mid$Y$_{n}$)$\mid$Y$_{1}\ldots$ & $\lambda$ y$_{n}\ldots$,y$_{1}$ [X(y$_{1}\ldots$,y$_{n}$)] $\Rightarrow$\  $\lambda$ y$_{1}$,y$_{n}\ldots$ [X(y$_{1}\ldots$,y$_{n}$)]\\[0.2cm]
\hline
\hline
\end{tabular}
\caption{GCG Rule Schemata}
\label{rules}
\end{center}
\end{figure*}

\begin{figure}
\begin{center}
\begin{tabular}{lll}
Kim    & loves                     & Sandy \\
NP     & (S$\backslash$NP)/NP       & NP \\
kim'   & $\lambda$ y,x [love$'$(x y)]  & sandy$'$\\
       & --------------P               &\\
       & (S/NP)$\backslash$NP       &\\
       & $\lambda$ x,y [love$'$(x y)]  &\\
\multicolumn{2}{l}{-----------------------------BA} &\\
S/NP & &\\
$\lambda$ y [love$'$(kim$'$ y)] & &\\
\multicolumn{3}{l}{---------------------------------------------------FA}\\
S & &\\
love$'$(kim$'$ sandy$'$) &&
\end{tabular} 
\end{center}
\caption{GCG Derivation for {\it Kim loves Sandy}}
\label{p-ba}
\end{figure}
GCG as presented is inadequate as an account of UG or of any
individual grammar. In particular, the definition of atomic categories
needs extending to deal with featural variation (e.g. Bouma and van
Noord, 1994), and the rule schemata, especially composition and weak
permutation, must be restricted in various parametric ways so that
overgeneration is prevented for specific languages.
Nevertheless, GCG does represent a plausible kernel of UG;
Hoffman (1995, 1996) explores the descriptive power of a very similar
system, in which generalized weak permutation is not required because
functor arguments are interpreted as multisets. She demonstrates that
this system can handle (long-distance) scrambling elegantly and
generates mildly context-sensitive languages (Joshi {\it et al}, 1991).

The relationship between GCG as a theory of UG (GCUG) and as a the
specification of a particular grammar is captured by embedding the
theory in a default inheritance hierarchy. This is represented as a
lattice of typed default feature structures (TDFSs) representing
subsumption and default inheritance relationships (Lascarides {\it et
  al}, 1996; Lascarides and Copestake, 1996). The lattice defines
intensionally the set of possible categories and rule schemata via
type declarations on nodes.  For example, an intransitive verb might
be treated as a subtype of verb, inheriting subject directionality by
default from a type {\bf gendir} (for general direction). For English,
{\bf gendir} is default {\bf right} but the node of the (intransitive)
functor category, where the directionality of subject arguments is
specified, overrides this to {\bf left}, reflecting the fact that
English is predominantly right-branching, though subjects appear to
the left of the verb.  A transitive verb would inherit structure from
the type for intransitive verbs and an extra NP argument with default
directionality specified by {\bf gendir}, and so forth.\footnote{Bouma
  and van Noord (1994) and others demonstrate that CGs can be embedded
  in a constraint-based representation.  Briscoe (1997a,b) gives
  further details of the encoding of GCG in TDFSs.}

For the purposes of the evolutionary simulation described in \S3,
GC(U)Gs are represented as a sequence of {\it p-settings} (where
{\it p} denotes principles or parameters) based on a flat (ternary)
sequential encoding of such default inheritance lattices.  The
inheritance hierarchy provides a partial ordering on parameters, which
is exploited in the learning procedure. For example, the atomic
categories, {\bf N}, {\bf NP} and {\bf S} are each represented by a
parameter encoding the presence/absence or lack of specification
(T/F/?)  of the category in the (U)G.  Since they will be unordered in the
lattice their ordering in the sequential coding is arbitrary.
However, the ordering of the directional types {\bf gendir} and {\bf
subjdir} (with values L/R) is significant as the latter is a more
specific type. The distinctions between absolute, default or unset
specifications also form part of the encoding (A/D/?).
Figure~\ref{binencoding} shows several equivalent and equally correct
sequential encodings of the fragment of the English type system
outlined above.
\begin{figure*}
\begin{center}
\begin{tabular}{|llllll|}
\hline
{\bf NP} & {\bf N} & {\bf S} & {\bf gen-dir} & {\bf subj-dir} & {\bf applic}\\
A T     & A T     & A T     & D R           & D L             & D T\\
\hline
{\bf NP} & {\bf gendir} & {\bf applic} & {\bf S} & {\bf N} & {\bf subj-dir} \\
A T     & D R            & D T           & A T     & A T     & D L \\
\hline
{\bf applic} & {\bf NP} & {\bf N} & {\bf gen-dir} & {\bf subj-dir} & {\bf S} \\
D T          & A T     & A T     & D R             & D L          & A T \\
\hline 
\ldots &&&&& \\
\hline
\end{tabular}
\end{center}
\caption{Sequential encodings of the grammar fragment}
\label{binencoding}
\end{figure*}

A set of grammars based on typological distinctions defined by basic
constituent order (e.g. Greenberg, 1966; Hawkins, 1994) was
constructed as a (partial) GCUG with independently varying
binary-valued parameters. The eight basic language families are
defined in terms of the unmarked order of verb (V), subject (S) and
objects (O) in clauses.  Languages within families further specify the
order of modifiers and specifiers in phrases, the order of adpositions
and further phrasal-level ordering parameters. Figure~\ref{lgset}
list the language-specific ordering parameters used to define the
full set of grammars in (partial) order of generality, and
gives examples of settings based on familiar languages such as
``English'', ``German'' and ``Japanese''.\footnote{Throughout
  double quotes around language names are used as
  convenient mnemonics for familiar combinations
  of parameters. Since not all aspects of 
  these actual languages are represented in the grammars,
  conclusions about actual languages must be made with care.}
\begin{figure*}
\begin{center}
\begin{tabular}{|l|lllllllllll|}
\hline 
& {\bf gen} & {\bf v1} & {\bf n} & {\bf subj} & {\bf obj} & {\bf v2}
& {\bf mod} & {\bf spec} & {\bf relcl} & {\bf adpos} & {\bf compl}\\
\hline
Engl & R   & F        & R       & L           & R        & F
& R         & R          & R           & R           & R\\
Ger  & R   & F        & R       & L           & L        & T
& R         & R          & R           & R           & R\\
Jap  & L   & F        & L       & L           & L        & F
& L         & L          & L           & L           & ?\\
\hline
\end{tabular}
\end{center}
\caption{The Grammar Set -- Ordering Parameters}
\label{lgset}
\end{figure*}
``English'' defines an SVO language, with prepositions in which
specifiers, complementizers and some modifiers precede heads of
phrases. There are other grammars in the SVO family in which all
modifers follow heads, there are postpositions, and so forth. Not all
combinations of parameter settings correspond to attested languages
and one entire language family (OVS) is unattested. ``Japanese'' is an
SOV language with postpositions in which specifiers and modifiers
follow heads.  There are other languages in the SOV family with less
consistent left-branching syntax in which specifiers and/or modifiers
precede phrasal heads, some of which are attested. ``German'' is a
more complex SOV language in which the parameter verb-second (v2)
ensures that the surface order in main clauses is usually
SVO.\footnote{Representation of the v1/v2 parameter(s) in terms of a
  type constraint determining allowable functor categories is
  discussed in more detail in Briscoe (1997b).}

There are 20 p-settings which determine the rule schemata available,
the atomic category set, and so forth. In all, this CGUG defines
just under 300 grammars.  Not all of the resulting languages are
(stringset) distinct and some are proper subsets of other languages.
``English'' without the rule of permutation results in a
stringset-identical language, but the grammar assigns different
derivations to some strings, though the associated logical forms are
identical.  ``English'' without composition results in a subset
language. Some combinations of p-settings result in `impossible'
grammars (or UGs). Others yield equivalent grammars, for example,
different combinations of default settings (for types and their
subtypes) can define an identical category set.

The grammars defined generate (usually infinite) stringsets of lexical
syntactic categories. These strings are sentence types since each is
equivalent to a finite set of grammatical sentences formed by
selecting a lexical instance of each lexical category.
Languages are represented as a finite subset of
sentence types generated by the associated grammar. These represent a
sample of degree-1 learning triggers for the language (e.g.
Lightfoot, 1991).  Subset languages are represented by 3-9 sentence
types and `full' languages by 12 sentence types. The constructions
exemplified by each sentence type and their length are equivalent
across all the languages defined by the grammar set, but the sequences
of lexical categories can differ. For example, two SOV language
renditions of {\it The man who Bill likes gave Fred a present}, one
with premodifying and the other postmodifying relative clauses, both
with a relative pronoun at the right boundary of the relative clause,
are shown below with the differing category highlighted.\\[3mm]
Bill likes who the-man a-present Fred gave\\
NP$_{s}$ (S$\backslash$NP$_{s}$)$\backslash$NP$_{o}$
Rc$\backslash$(S$\backslash$NP$_{o}$) {\bf NP$_{s}\backslash$Rc}
NP$_{o2}$ NP$_{o1}$
((S$\backslash$NP$_{s}$)$\backslash$NP$_{o2}$)$\backslash$NP$_{o1}$\\[2mm]
The-man Bill likes who a-present Fred gave\\ {\bf NP$_{s}$/Rc}
NP$_{s}$ (S$\backslash$NP$_{s}$)$\backslash$NP$_{o}$
Rc$\backslash$(S$\backslash$NP$_{o}$) NP$_{o2}$ NP$_{o1}$
((S$\backslash$NP$_{s}$)$\backslash$NP$_{o2}$)$\backslash$NP$_{o1}$\\[3mm]

\subsection{The Parser}

The parser is a deterministic, bounded-context stack-based
shift-reduce algorithm. The parser operates on two data structures, an
input buffer or queue, and a stack or push down store.  The algorithm
for the parser working with a GCG which includes application,
composition and permutation is given in Figure~\ref{parsalg}.
\begin{figure}
\begin{enumerate}

\item {\bf The Reduce Step}: if the top 2 cells of the stack are occupied,\\
then try\\
a) Application, if match, then apply and goto 1), else b),\\
b) Combination, if match then apply and goto 1), else c),\\
c) Permutation, if match then apply and goto 1), else goto 2)

\item {\bf The Shift Step}: if the first cell of the Input Buffer is occupied,\\
then pop it and move it onto the Stack together with its associated lexical
syntactic category and goto 1),\\ 
else goto 3)

\item {\bf The Halt Step}: if only the top cell of the Stack is occupied by a
constituent of category S,\\ 
then return Success,\\ 
else return Fail

\end{enumerate}

{\bf The Match and Apply operation}: if a binary rule schema matches the
categories of the top 2 cells of the Stack, then they are popped from
the Stack and the new category formed by applying the rule schema is
pushed onto the Stack.

\vspace*{3mm}

{\bf The Permutation operation}: each time step 1c) is visited during
the Reduce step, permutation is applied to one of the categories in
the top 2 cells of the Stack until all possible permutations of the 2
categories have been tried using the binary rules. The number of
possible permutation operations is finite and bounded by the maximum
number of arguments of any functor category in the grammar.

\caption{The Parsing Algorithm}
\label{parsalg}
\end{figure}
This algorithm finds the most left-branching derivation for a sentence
type because Reduce is ordered before Shift.  The category sequences
representing the sentence types in the data for the entire language
set are designed to be unambiguous relative to this `greedy',
deterministic algorithm, so it will always assign the appropriate
logical form to each sentence type. However, there are frequently
alternative less left-branching derivations of the same logical form.

The parser is augmented with an algorithm which computes working
memory load during an analysis (e.g. Baddeley, 1992).  Limitations of
working memory are modelled in the parser by associating a cost with
each stack cell occupied during each step of a derivation, and recency
and depth of processing effects are modelled by resetting this cost
each time a reduction occurs: the working memory load (WML) algorithm
is given in Figure~\ref{wml}.  Figure~\ref{Kim2} gives the
right-branching derivation for {\it Kim loves Sandy}, found by the
parser utilising a grammar without permutation. The WML at each step
is shown for this derivation. The overall WML (16) is higher than for
the left-branching derivation (9).
\begin{figure}

After each parse step (Shift, Reduce, Halt (see Fig~\ref{parsalg}):
\begin{enumerate}

\item Assign any new Stack entry in the top cell (introduced by Shift
or Reduce) a WML value of 0

\item Increment every Stack cell's WML value by 1

\item Push the sum of the WML values of each Stack cell onto the
WML-record

\end{enumerate}

When the parser halts, return the sum of the WML-record gives the
total WML for a derivation

\caption{The WML Algorithm}
\label{wml}
\end{figure}
\begin{figure*}
\begin{center}
\begin{tabular}{|lllll|}
\hline
Stack                            & Input Buffer       & Operation & Step & WML\\
\hline
                                 & Kim loves Sandy    &           & 0 & 0 \\
\hline
Kim:NP:kim$'$                    & loves Sandy        & Shift     & 1 & 1\\
\hline
loves:(S$\backslash$NP)/NP:$\lambda$ y,x(love$'$ x, y) & Sandy   & Shift     & 2 & 3\\
Kim:NP:kim$'$                    & & & &\\
\hline
Sandy:NP:sandy$'$                           &         & Shift     & 3 & 6\\
loves:(S$\backslash$NP)/NP:$\lambda$ y,x(love$'$ x, y) & & & &\\
Kim:NP:kim$'$                    & & & &\\
\hline
loves Sandy:S/NP:$\lambda$ x(love$'$ x, sandy$'$) &    & Reduce (A) & 4 & 5\\
Kim:NP:kim$'$                    & & & &\\
\hline
Kim loves Sandy:S:(love$'$ kim$'$, sandy$'$)  &       & Reduce (A) & 5 & 1\\
\hline
\end{tabular}
\end{center}
\caption{WML for {\it Kim loves Sandy}}
\label{Kim2}
\end{figure*}

The WML algorithm ranks sentence types, and thus indirectly languages,
by parsing each sentence type from the exemplifying data with the
associated grammar and then taking the mean of the WML obtained for
these sentence types.  ``English'' with Permutation has a lower mean
WML than ``English'' without Permutation, though they are
stringset-identical, whilst a hypothetical mixture of ``Japanese'' SOV
clausal order with ``English'' phrasal syntax has a mean WML which is
25\% worse than that for ``English''. The WML algorithm is in accord
with existing (psycholinguistically-motivated) theories of parsing
complexity (e.g.  Gibson, 1991; Hawkins, 1994; Rambow and Joshi,
1994).

\subsection{The Parameter Setting Algorithm}

The parameter setting algorithm is an extension of Gibson and Wexler's
(1994) Trigger Learning Algorithm (TLA) to take account of the
inheritance-based partial ordering and the role of memory in learning.
The TLA is error-driven -- parameter settings are altered in constrained
ways when a learner cannot parse trigger input.  Trigger input is
defined as primary linguistic data which, because of its structure or
context of use, is determinately unparsable with the correct
interpretation (e.g. Lightfoot, 1991). In this model, the issue of
ambiguity and triggers does not arise because all sentence types are
treated as triggers represented by p-setting schemata. The
TLA is memoryless in the sense that a history of parameter
(re)settings is not maintained, in principle, allowing the learner to
revisit previous hypotheses. This is what allows Niyogi and Berwick
(1995) to formalize parameter setting as a Markov process.  However,
as Brent (1996) argues, the psychological plausibility of this
algorithm is doubtful -- there is no evidence that children (randomly)
move between neighbouring grammars along paths that revisit previous
hypotheses.  Therefore, each parameter can only be reset once
during the learning process. Each step for a learner can be defined in
terms of three functions: {\sc p-setting}, {\sc grammar} and {\sc
  parser}, as:\\[3mm]
{\sc parser}$_{i}$({\sc grammar}$_{i}$({\sc p-setting}$_{i}$(Sentence$_{j}$)))\\[3mm] 
A p-setting defines a grammar which in turn defines a parser (where
the subscripts indicate the output of each function given the previous
trigger). A
parameter is updated on parse failure and, if this results in a parse,
the new setting is retained. The algorithm is summarized in
Figure~\ref{learnalg}.
\begin{figure}[tp]

Data: \{S$_{1}$, S$_{2}$, $\ldots$\ S$_{n}$\}\\

unless\\
{\sc parser}$_{i}$({\sc grammar}$_{i}$({\sc p-setting}$_{i}$(S$_{j}$))) = Success\\
then\\
\hspace*{.3cm}p-setting$_{j}$ = {\sc update}(p-setting$_{i}$)\\
\hspace*{.3cm}unless\\
\hspace*{.3cm}{\sc parser}$_{j}$({\sc grammar}$_{j}$({\sc p-setting}$_{j}$(S$_{j}$))) = Success\\
\hspace*{.3cm}then\\
\hspace*{.6cm}RETURN p-settings$_{i}$\\
\hspace*{.3cm}else\\
\hspace*{.6cm}RETURN p-settings$_{j}$\\

Update:\\
Reset the first (most general) default or unset parameter in a
left-to-right search of the p-set according to the following
table:\\

\begin{tabular}{|llll|}
\hline
Input:  & D 1 & D 0 & ? ?\\
Output: & R 0 & R 1 & ? 1/0 (random)\\
\hline
\end{tabular}
(where 1 = T/L and 0 = F/R)
\caption{The Learning Algorithm}
\label{learnalg}
\end{figure}
Working memory grows through childhood (e.g. Baddeley, 1992), and this
may assist learning by ensuring that trigger sentences gradually
increase in complexity through the acquisition period (e.g.  Elman,
1993) by forcing the learner to ignore more complex potential triggers
that occur early in the learning process.  The WML of a sentence type
can be used to determine whether it can function as a trigger at a
particular stage in learning.

\section{The Simulation Model}

The computational simulation supports the evolution of a population of
Language Agents (LAgts), similar to Holland's (1993) Echo agents.
LAgts generate and parse sentences compatible with their current
p-setting. They participate in linguistic interactions which are
successful if their p-settings are compatible.  The relative fitness
of a LAgt is a function of the proportion of its linguistic
interactions which have been successful, the expressivity of the
language(s) spoken, and, optionally, of the mean WML for parsing
during a cycle of interactions. An interaction cycle consists of a
prespecified number of individual random interactions between LAgts,
with generating and parsing agents also selected randomly.  LAgts
which have a history of mutually successful interaction and high
fitness can `reproduce'.  A LAgt can `live' for up to ten interaction
cycles, but may `die' earlier if its fitness is relatively low.  It is
possible for a population to become extinct (for example, if all the
initial LAgts go through ten interaction cycles without any successful
interaction occurring), and successful populations tend to grow at a
modest rate (to ensure a reasonable proportion of adult speakers is
always present).  LAgts learn during a critical period from ages 1-3
and reproduce from 4-10, parsing and/or generating any language learnt
throughout their life.

During learning a LAgt can reset genuine parameters which either were
unset or had default settings `at birth'. However, p-settings with an
absolute value (principles) cannot be altered during the lifetime of
an LAgt. Successful LAgts reproduce at the end of
interaction cycles by one-point crossover of (and, optionally, single
point mutation of) their {\it initial} p-settings, ensuring
neo-Darwinian rather than Lamarckian inheritance.  The encoding of
p-settings allows the deterministic recovery of the initial setting.
Fitness-based reproduction ensures that successful and somewhat
compatible p-settings are preserved in the population and randomly
sampled in the search for better versions of universal grammar,
including better initial settings of genuine parameters. Thus,
although the learning algorithm {\it per se} is fixed, a range of
alternative learning procedures can be explored based on the
definition of the inital set of parameters and their initial settings.
Figure~\ref{simvar} summarizes crucial options in the simulation
giving the values used in the experiments reported in \S4 and
Figure~\ref{fitness} shows the fitness functions.
\begin{figure}
\begin{tabular}{lll}
{\bf Variables} & \multicolumn{2}{c}{\bf Typical Values} \\
Population Size & & 32\\
Interaction Cycle & 2K Interactions &\\
Simulation Run & 50 Cycles &\\
Crossover Probability & & 0.9\\
Mutation Probability & & 0\\                  
Learning & memory limited & yes\\
         & critical period & yes
\end{tabular}
\caption{The Simulation Options}
\label{simvar}
\end{figure}
\begin{figure}
(Cost/Benefits per sentence (1--6); summed for each LAgt at end of an
interaction cycle and used to calculate fitness functions (7--8)):\\
\begin{enumerate}
\item Generate cost: 1 (GC)
\item Parse cost: 1 (PC)
\item Generate subset language cost: 1 (GSC)
\item Parse failure cost: 1 (PF)
\item Parse memory cost: WML(st)
\item Interaction success benefit: 1 (SI)
\item Fitness(WML): $\frac{SI}{GC+PC} \times \frac{GC}{GC+GSC}
  \times \frac{1}{(\frac{WML}{PC-PF})}$
\item Fitness($\neg$WML): $\frac{SI}{GC+PC} \times \frac{GC}{GC+GSC}$
\end{enumerate}
\caption{Fitness Functions}
\label{fitness}
\end{figure}

\section{Experimental Results}

\subsection{Effectiveness of Learning Procedures}

Two learning procedures were predefined -- a default learner and an
unset learner. These LAgts were initialized with p-settings consistent
with a minimal inherited CGUG consisting of application with NP and S
atomic categories. All the remaining p-settings were genuine
parameters for both learners. The unset learner was initialized with
all unset, whilst the default learner had default settings for the
parameters {\bf gendir} and {\bf subjdir} and {\bf argorder} which
specify a minimal SVO right-branching grammar, as well as default
(off) settings for {\bf comp} and {\bf perm} which determine the
availability of Composition and Permutation, respectively. The unset
learner represents a `pure' principles-and-parameters learner. The
default learner is modelled on Bickerton's bioprogram learner.

Each learner was tested against an adult LAgt initialized to generate
one of seven full languages in the set which are close to an attested
language; namely, ``English'' (SVO, predominantly right-branching),
``Welsh'' (SVOv1, mixed order), ``Malagasy'' (VOS, right-branching),
``Tagalog'' (VSO, right-branching), ``Japanese'' (SOV,
left-branching), ``German'' (SOVv2, predominantly right-branching),
``Hixkaryana'' (OVS, mixed order), and an unattested full OSV language
with left-branching syntax. In these tests, a single learner
interacted with a single adult.  After every ten interactions, in which
the adult randomly generated a sentence type and the learner
attempted to parse and learn from it, the state of the
learner's p-settings was examined to determine whether the learner had
converged on the same grammar as the adult.  Table~\ref{learneffect}
shows the number of such interaction cycles (i.e.  the number of input
sentences to within ten) required by each type of learner to converge
on each of the eight languages.
\begin{table*}
\begin{center}
\begin{tabular}{|l|llllllll|}
\hline
{\bf Learner} & \multicolumn{8}{c|}{{\bf Language}} \\
\hline
        & SVO & SVOv1 & VOS & VSO & SOV & SOVv2 & OVS & OSV\\
\hline
Unset   & 60  & 80    & 70  & 80  & 70  & 70    & 70  & 70\\
Default & 60  & 60    & 60  & 60  & 60  & 60    & 80  & 70\\
\hline
\end{tabular}
\end{center}
\caption{Effectiveness of Two Learning Procedures}
\label{learneffect}
\end{table*}
These figures are each calculated from 100 trials to a 1\% error rate;
they suggest that, in general, the default learner is more effective
than the unset learner. However, for the OVS language (OVS languages
represent 1.24\% of the world's languages, Tomlin, 1986), and for the
unattested OSV language, the default (SVO) learner is less effective.
So, there are at least two learning procedures in the space defined by
the model which can converge with some presentation orders on some of
the grammars in this set. Stronger conclusions require either
exhaustive experimentation or theoretical analysis of the model of the
type undertaken by Gibson and Wexler (1994) and Niyogi and Berwick
(1995).

\subsection{Evolution of Learning Procedures}

In order to test the preference for default versus unset parameters
under different conditions, the five parameters which define
the difference between the two learning procedures were tracked
through another series of 50 cycle runs initialized with either 16
default learning adult speakers and 16 unset learning adult speakers,
with or without memory-limitations during learning and
parsing, speaking one of the eight languages described above. Each
condition was run ten times.  In the memory limited runs, default
parameters came to dominate some but not all populations. 
\hidden{
For example,
Figure~\ref{vos-mem-du1} shows a typical plot of the percentage of the
population with default settings for these five parameters for a run
initialized with ``Malagasy'' VOS speakers and memory limited learning
and parsing.
\begin{figure}
\begin{center}
\includegraphics{vos-defs1.ps}
\caption{Percentage of parameters types with WML}
\label{vos-mem-du1}
\end{center}
\end{figure}
At the start of this run, the percentage of unset and default
parameters is identical. However, as differential reproduction sets
in, we can see that there is a clear increase in the proportion of
default parameters in the population.}
In a few runs all unset
parameters disappeared altogether. In all runs with populations
initialized to speak ``English'' (SVO) or ``Malagasy'' (VOS) the
preference for default settings was 100\%. In 8 runs with ``Tagalog''
(VSO) the same preference emerged, in one there was a preference for
unset parameters and in the other no clear preference.  However, for
the remaining five languages there was no strong preference.

The results for the runs without memory limitations are different,
with an increased preference for unset parameters across all languages
but no clear 100\% preference for any individual language.
Table~\ref{udn} shows the pattern of preferences which emerged across
160 runs and how this was affected by the presence or absence of
memory limitations.
\begin{table}
\begin{center}
\begin{tabular}{|r|lll|}
\hline
           & Unset & Default & None \\
\hline
WML        & 15    & 39      & 26 \\
$\neg$WML  & 34    & 17      & 29 \\
\hline
\end{tabular}
\end{center}
\caption{Overall preferences for parameter types}
\label{udn}
\end{table}

To test whether it was memory limitations during learning or during
parsing which were affecting the results, another series of runs for
``English'' was performed with either memory limitations during
learning but not parsing enabled, or vice versa. Memory limitations
during learning are creating the bulk of the preference for a default
learner, though there appears to be an additive effect.  In seven of the
ten runs with memory limitations only in learning, a clear preference
for default learners emerged. In five of the runs with memory limitations
only in parsing there appeared to be a slight preference for defaults
emerging. Default learners may have a fitness advantage when the number of
interactions required to learn successfully is greater because they
will tend to converge faster, at least to a subset language. This will
tend to increase their fitness over unset learners who do not speak
any language until further into the learning period.

\begin{figure}[t]
\begin{center}
\includegraphics{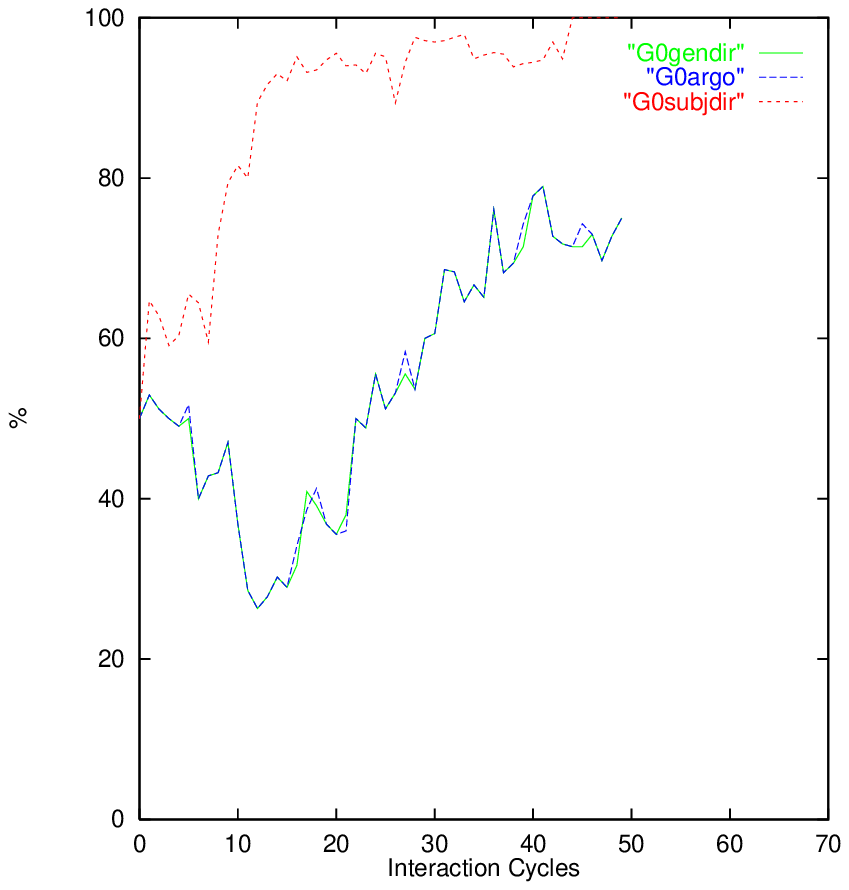}
\caption{Percentage of each default ordering parameter}
\label{vos-subjdir}
\end{center}
\end{figure}

The precise linguistic environment of adaptation determines the
initial values of default parameters which evolve. For example, in the
runs initialized with 16 unset learning ``Malagasy'' VOS adults and 16
default (SVO) learning VOS adults, the learning procedure which
dominated the population was a variant VOS default learner in which
the value for {\bf subjdir} was reversed to reflect the position of
the subject in this language.  In some of these runs, the entire
population evolved a default {\bf subjdir} `right' setting, though
some LAgts always retained unset settings for the other two ordering
parameters, {\bf gendir} and {\bf argo}, as is illustrated in
Figure~\ref{vos-subjdir}. This suggests that if the human language
faculty has evolved to be a right-branching SVO default learner, then
the environment of linguistic adaptation must have contained a
dominant language fully compatible with this (minimal) grammar.

\subsection{Emergence of Language and Learners}

To explore the emergence and persistence of structured language, and
consequently the emergence of effective learners, (pseudo) random
initialization was used. A series of simulation runs of 500 cycles
were performed with random initialization of 32 LAgts' p-settings for
any combination of p-setting values, with a probability of 0.25 that a
setting would be an absolute principle, and 0.75 a parameter with
unbiased allocation for default or unset parameters and for values of
all settings. All LAgts were initialized to be age 1 with a critical
period of 3 interaction cycles of 2000 random interactions for
learning, a maximum age of 10, and the ability to reproduce by
crossover (0.9 probability) and mutation (0.01 probability) from 4-10.
In around 5\% of the runs, language(s) emerged and persisted to the
end of the run.

\begin{figure}[t]
\begin{center}
\includegraphics{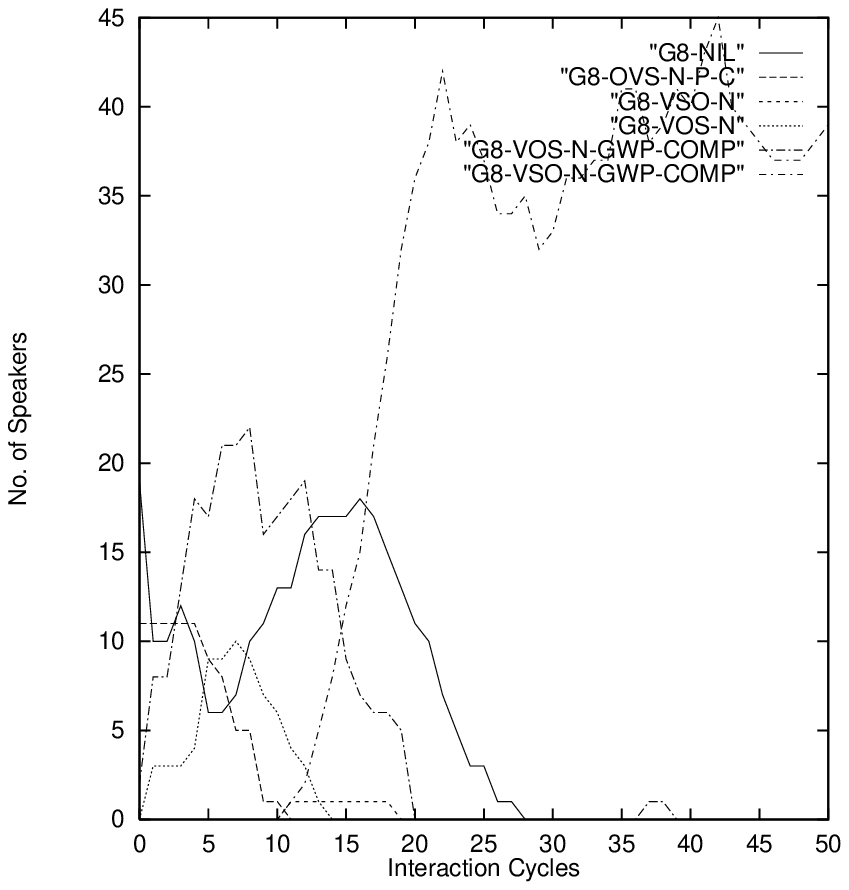}
\end{center}
\caption{Emergence of language(s)}
\label{emerg-lgs}
\end{figure}

Languages with close to optimal WML scores typically came to dominate
the population quite rapidly.  However, sometimes sub-optimal
languages were initially selected and occasionally these persisted
despite the later appearance of a more optimal language, but with few
speakers.  Typically, a minimal subset language dominated -- although
full and intermediate languages did appear briefly, they did not
survive against less expressive subset languages with a lower mean
WML. Figure~\ref{emerg-lgs} is a typical plot of the
emergence (and extinction) of languages in one of these runs.
In this run, around 10 of the initial population converged on a
minimal OVS language and 3 others on a VOS language. The latter is
more optimal with respect to WML and both are of equal expressivity
so, as expected, the VOS language acquired more speakers over the next
few cycles. A few speakers also converged on VOS-N, a more expressive
but higher WML extension of VSO-N-GWP-COMP. However, neither
this nor the OVS language survived beyond cycle 14. Instead a VSO
language emerged at cycle 10, which has the same minimal expressivity
of the VOS language but a lower WML (by virtue of placing the subject
before the object) and this language dominated rapidly and eclipsed
all others by cycle 40.

In all these runs, the population settled on subset languages of low
expressivity, whilst the percentage of absolute principles and default
parameters increased relative to that of unset parameters (mean \%
change from beginning to end of runs: +4.7, +1.5 and -6.2,
respectively). So a second identical set of ten was undertaken, except
that the initial population now contained two SOV-V2 ``German''
speaking unset learner LAgts. In seven of these runs, the population
fixed on a full SOV-V2 language, in two on the intermediate subset
language SOV-V2-N, and in one on the minimal subset language
SOV-V2-N-GWP-COMP.  These runs suggest that if a full language defines
the environment of adaptation then a population of randomly
initialized LAgts is more likely to converge on a (related) full
language. Thus, although the simulation does not model the development
of expressivity well, it does appear that it can model the emergence
of effective learning procedures for (some) full languages. The
pattern of language emergence and extinction followed that of the
previous series of runs: lower mean WML languages were selected from
those that emerged during the run.  However, often the initial optimal
SVO-V2 itself was lost before enough LAgts evolved capable of learning
this language. In these runs, changes in the percentages of absolute,
default or unset p-settings in the population show a marked
difference: the mean number of absolute principles declined by 6.1\%
and unset parameters by 17.8\%, so the number of default parameters
rose by 23.9\% on average between the beginning and end of the 10
runs. This may reflect the more complex linguistic environment in
which (incorrect) absolute settings are more likely to handicap,
rather than simply be irrelevant to, the performance of the LAgt.

\section{Conclusions}

Partially ordering the updating of parameters can result in
(experimentally) effective learners with a more complex parameter
system than that studied previously. Experimental comparison of the
default (SVO) learner and the unset learner suggests that the default
learner is more efficient on typologically more common constituent
orders. Evolutionary simulation predicts that a learner with default
parameters is likely to emerge, though this is dependent both on the
type of language spoken and the presence of memory limitations during
learning and parsing.  Moreover, a SVO bioprogram learner is only
likely to evolve if the environment contains a dominant SVO language.

The evolution of a bioprogram learner is a manifestation of the
Baldwin Effect (Baldwin, 1896) -- genetic assimilation of aspects of
the linguistic environment during the period of evolutionary
adaptation of the language learning procedure. In the case of grammar
learning this is a co-evolutionary process in which languages (and
their associated grammars) are also undergoing selection. The WML
account of parsing complexity predicts that a right-branching SVO
language would be a near optimal selection at a stage in grammatical
development when complex rules of reordering such as extraposition,
scrambling or mixed order strategies such as v1 and v2 had not
evolved. Briscoe (1997a) reports further experiments which demonstrate
language selection in the model.

Though, simulation can expose likely evolutionary pathways under
varying conditions, these might have been blocked by
accidental factors, such as genetic drift or bottlenecks, causing
premature fixation of alleles in the genotype (roughly corresponding
to certain p-setting values).  The value of the simulation is to,
firstly, show that a bioprogram learner could have emerged via
adaptation, and secondly, to clarify experimentally the precise
conditions required for its emergence. Since in many cases these
conditions will include the presence of constraints (working memory
limitations, expressivity, the learning algorithm etc.) which will
remain causally manifest, further testing of any conclusions drawn
must concentrate on demonstrating the accuracy of the assumptions made
about such constraints. Briscoe (1997b) evaluates the psychological
plausibility of the account of parsing and working memory.

\section*{References}

\newcommand{\book}[4]{\item #1 (#4) {\it #2,} #3.}
\newcommand{\booknoauthor}[3]{\item #1 (#3) {\it #2,}}
\newcommand{\forthedbook}[4]{\item #1 (eds.) (#4, forthcoming) {\it #2,} #3.}
\newcommand{\unpub}[4]{\item #1 (#4) {\it #2,} #3.}
\newcommand{\bookart}[7]{\item #1 (#7) `#2' in #5 (ed.), {\it #4,} #6, pp.~#3.}
\newcommand{\bookartnopp}[6]{\item #1 (#6) `#2' in #4 (ed.), {\it #3,} #5.}
\newcommand{\forthart}[6]{\item #1 (#6, forthcoming) `#2' in #4 (eds.),
{\it #3,} #5.}
\newcommand{\journart}[6]{\item #1 (#6) `#2', {\it #3,} {\it vol.#4,}
#5.}
\newcommand{\journartnopp}[5]{\item #1 (#5) `#2', {\it #3,} {\it vol.#4,}}
\newcommand{\forthjournart}[4]{\item #1 (#4) `#2', {\it #3,}}
\newcommand{\procart}[6]{\item #1 (#6) `#2', {\it Proceedings of the
#3,} #4, pp.~#5.}
\newcommand{\forthprocart}[5]{\item #1 (#5) `#2', 
to appear in {\it Proceedings of the #3,} #4}
\newcommand{\procartnopp}[5]{\item #1 (#5) `#2', {\it Proceedings of the
#3,} #4.}

\begin{list}{}
   {\leftmargin 1.6em
    \itemindent -\leftmargin
    \itemsep 0pt plus 1pt
    \parsep 0pt plus 1pt}

\journart{Baddeley, A.}
     {Working Memory: the interface between memory and cognition}
     {J. of Cognitive Neuroscience}
     {4.3}
     {281--288}
     {1992}

\journart{Baldwin, J.M.}
         {A new factor in evolution}
         {American Naturalist}
         {30}
         {441--451}
         {1896}

\journart{Bickerton, D.}
         {The language bioprogram hypothesis}
         {The Behavioral and Brain Sciences}
         {7.2}
         {173--222}
         {1984}

\procart{Bouma, G. and van Noord, G}
        {Constraint-based categorial grammar}
        {32nd Assoc. for Computational Linguistics}
        {Las Cruces, NM}
        {147--154}
        {1994}

\journart{Brent, M.}
         {Advances in the computational study of language acquisition}
         {Cognition}
         {61}
         {1--38}
         {1996}

\forthjournart{Briscoe, E.J.}
      {Language Acquisition: the Bioprogram Hypothesis and the Baldwin Effect}
      {Language}
      {1997a, submitted}

\unpub{Briscoe, E.J.}
      {Working memory and its influence on the development of
human languages and the human language faculty}
      {University of Cambridge, Computer Laboratory, m.s.}
      {1997b, in prep.}

\book{Chomsky, N.}
     {Government and Binding}
     {Foris, Dordrecht}
     {1981}

\journart{Clark, R.}
         {The selection of syntactic knowledge}
         {Language Acquisition}
         {2.2}
         {83--149}
         {1992}

\journart{Elman, J.}
         {Learning and development in neural networks: the importance
         of starting small}
         {Cognition}
         {48}
         {71--99}
         {1993}

\book{Gibson, E.}
     {A Copmutational Theory of Human Linguistic Processing: Memory
     Limitations and Processing Breakdown}
     {Doctoral dissertation, Carnegie Mellon University}
     {1991}

\journart{Gibson, E. and Wexler, K.}
         {Triggers}
         {Linguistic Inquiry}
         {25.3}
         {407--454}
         {1994}

\bookart{Greenberg, J.}
        {Some universals of grammar with particular reference to the
order of meaningful elements} 
        {73--113}
        {Universals of Grammar}
        {J.~Greenberg}
        {MIT Press, Cambridge, Ma.}
        {1966}

\book{Hawkins, J.A.}
     {A Performance Theory of Order and Constituency}
     {Cambridge University Press, Cambridge}
     {1994}

\unpub{Hoffman, B.}
      {The Computational Analysis of the Syntax and Interpretation of
      `Free' Word Order in Turkish}
      {PhD dissertation, University of Pennsylvania}
      {1995}

\procartnopp{Hoffman, B.}
        {The formal properties of synchronous CCGs}
        {ESSLLI Formal Grammar Conference}
        {Prague}
        {1996}

\book{Holland, J.H.}
     {Echoing emergence: objectives, rough definitions and
speculations for echo-class models}
     {Santa Fe Institute, Technical Report 93-04-023}
     {1993}

\bookart{Joshi, A., Vijay-Shanker, K. and Weir, D.}
        {The convergence of mildly context-sensitive grammar
        formalisms}
        {31--82}
        {Foundational Issues in Natural Language Processing}
        {Sells, P., Shieber, S. and Wasow, T.}
        {MIT Press}
        {1991}

\journart{Lascarides, A., Briscoe E.J. , Copestake A.A and
Asher, N.}
              {Order-independent and persistent default unification}
              {Linguistics and Philosophy}
              {19.1}
              {1--89}
              {1995}

\forthjournart{Lascarides, A. and Copestake A.A.}
              {Order-independent typed default unification}
              {Computational Linguistics}
              {1996, submitted}

\book{Lightfoot, D.}
     {How to Set Parameters: Arguments from language Change}
     {MIT Press, Cambridge, Ma.}
     {1991}

\procartnopp{Niyogi, P. and Berwick, R.C.}
            {A markov language learning model for finite parameter
spaces}
            {33rd Annual Meeting of the Association for Computational
Linguistics} 
            {MIT, Cambridge, Ma.}
            {1995}

\bookart{Rambow, O. and Joshi, A.}
        {A processing model of free word order languages}
        {267--301}
        {Perspectives on Sentence Processing}
        {C.~Clifton, L.~Frazier and K.~Rayner}
        {Lawrence Erlbaum, Hillsdale, NJ.}
        {1994}

\bookart{Steedman, M.}
        {Combinators and grammars}
        {417--442}
        {Categorial Grammars and Natural Language Structures}
        {R.~Oehrle, E.~Bach and D.~Wheeler}
        {Reidel, Dordrecht}
        {1988}

\book{Tomlin, R.}
     {Basic Word Order: Functional Principles}
     {Routledge, London}
     {1986}

\book{Wood, M.M.}
     {Categorial Grammars}
     {Routledge, London}
     {1993}

\end{list}
\end{document}